\documentclass[letterpaper, 10 pt, conference]{ieeeconf}  

\IEEEoverridecommandlockouts                              

\usepackage{graphics} 
\usepackage{graphicx}
\usepackage{color}
\usepackage{amsmath} 
\usepackage{amssymb}  
\usepackage{amsthm}
\usepackage{mathtools}
\usepackage{verbatim}
\usepackage{soul}
\usepackage{multirow}
\usepackage[hidelinks]{hyperref}
\usepackage[bottom]{footmisc} 
\usepackage{flushend}

\usepackage{subcaption}
\usepackage[ruled, linesnumbered, vlined]{algorithm2e}
\SetAlFnt{\small}
\SetAlCapFnt{\small}
\SetAlCapNameFnt{\small}

\newtheorem{definition}{Definition}

\usepackage{upgreek}
\usepackage{bm}
\renewcommand{\b}[1]{\boldsymbol{#1}}

\newcommand{\h}[1]{{#1}}

\theoremstyle{remark}

\title{\LARGE \bf Globally Guided Trajectory Planning in Dynamic Environments} 
\author{Oscar de Groot, Laura Ferranti, Dariu Gavrila, Javier Alonso-Mora
\thanks{The authors are with the Dept. of Cognitive Robotics, TU Delft, 2628 CD Delft, The Netherlands. \texttt{Email: o.m.degroot@tudelft.nl}}
\thanks{This work received support from the Dutch Science Foundation NWO-TTW, within the SafeVRU project (nr. 14667) and Veni project HARMONIA (nr. 18165).}}%
\begin{document}
\maketitle
\thispagestyle{empty}
\pagestyle{empty}

\begin{abstract}
Navigating mobile robots through environments shared with humans is challenging. From the perspective of the robot, humans are dynamic obstacles that must be avoided. These obstacles make the collision-free space nonconvex, which leads to two distinct passing behaviors per obstacle (passing left or right). For local planners, such as receding-horizon trajectory optimization, each behavior presents a local optimum in which the planner can get stuck. This may result in slow or unsafe motion even when a better plan exists. 
In this work, we identify trajectories for multiple locally optimal driving behaviors, by considering their topology. This identification is made consistent over successive iterations by propagating the topology information. The most suitable high-level trajectory guides a local optimization-based planner, resulting in fast and safe motion plans. We validate the proposed planner on a mobile robot in simulation and real-world experiments.
\end{abstract}
\section{Introduction} \label{sec:introduction}
Mobile robots have the potential to automate logistic tasks ranging from indoor transportation tasks, as found in automated warehouses and hospitals, to outdoor transportation tasks, such as package delivery. One of the major challenges for mobile robots is to move safely among humans. 

Dynamic collision avoidance constraints are usually imposed on the motion of the robot, making its free configuration space nonconvex. In fact, each obstacle in a 2-D environment leads to at least two possible driving behaviors for the robot: passing left or right. 

Existing motion planners are either \textit{local} or \textit{global}. Local planners typically remain in the driving behavior that they are initialized with. This may lead to poor performance (e.g. long travel times) if a higher performance driving behavior exists, but is not explored. Because of the limited scope, these planners are typically fast and can consider detailed dynamic models. Global planners explore and find the optimal driving behavior and motion plan, but can suffer from high computation times when the robot dynamical model is considered.

\begin{figure}[t]
    \centering
    \begin{subfigure}[t]{0.18\textwidth}
        \centering
        \includegraphics*[width=\textwidth]{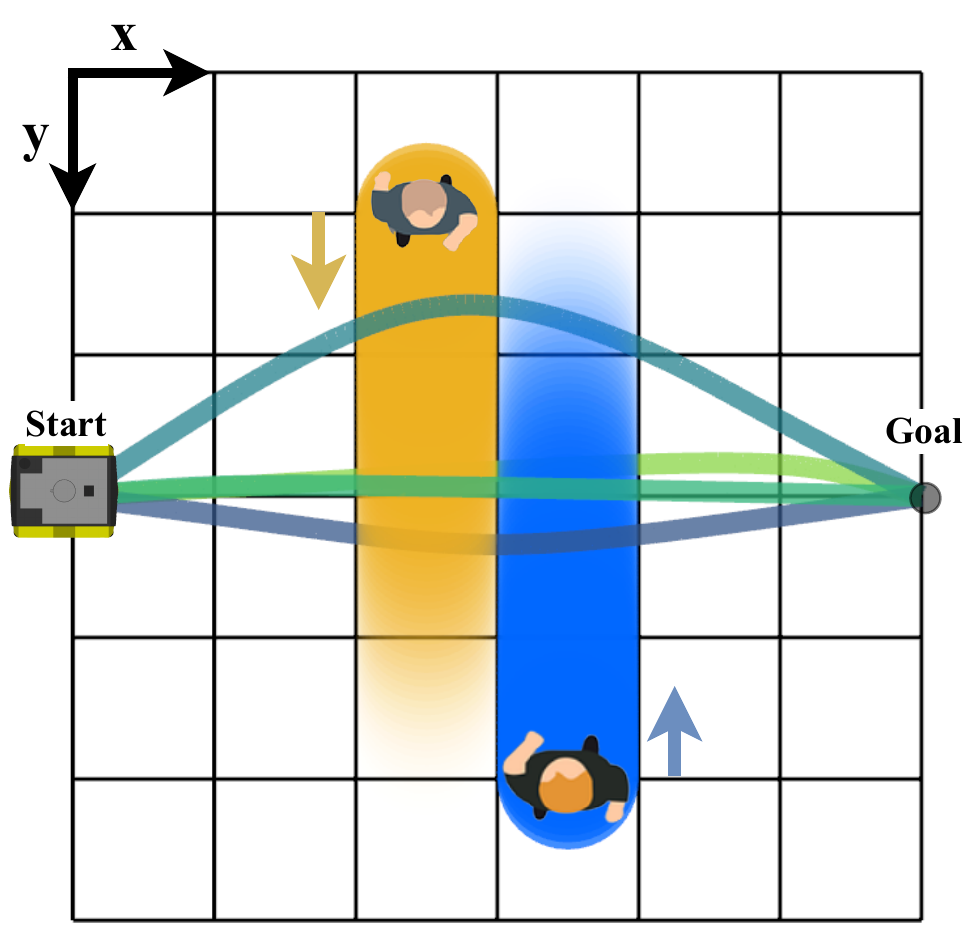}
        \caption{\h{Topview of a crossing scenario.}}
        \label{fig:3dplan-2d}
    \end{subfigure}
    \hspace{2pt}
    \begin{subfigure}[t]{0.20\textwidth}
        \centering
        \includegraphics*[width=\textwidth]{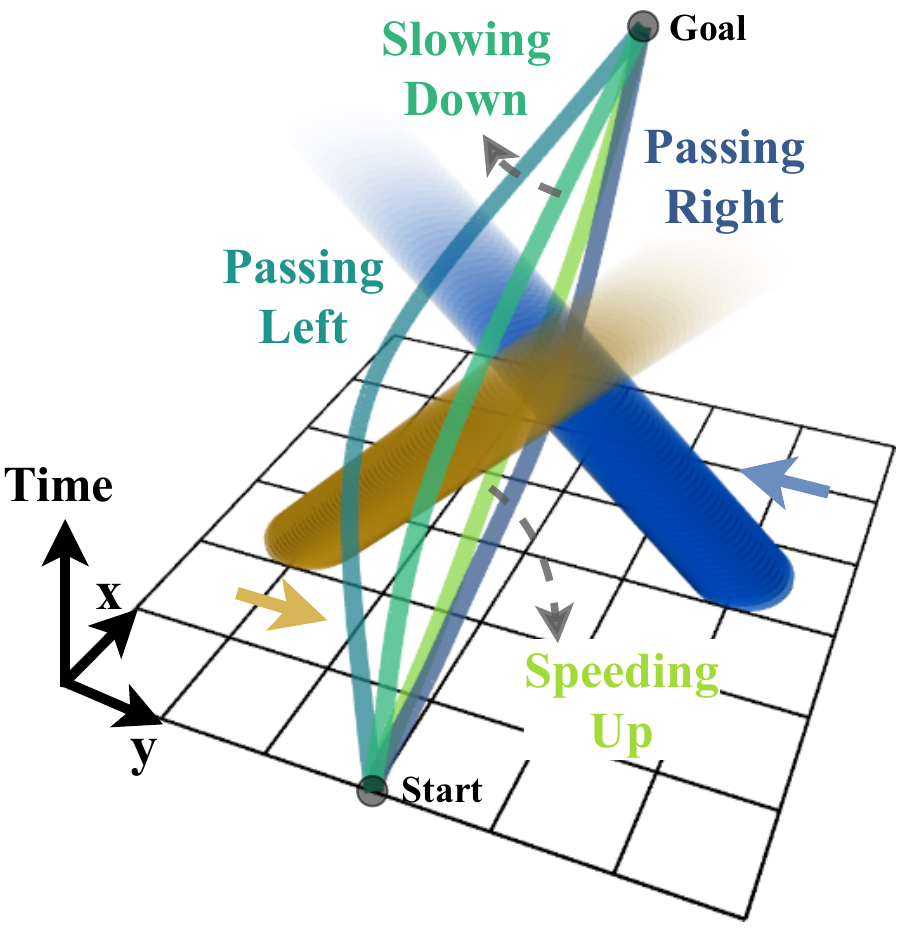}
        \caption{\h{Topologically distinct trajectories in the 3-D state space.}}
        \label{fig:3dplan-3d}
    \end{subfigure}
    \caption{An illustration of how topologically distinct guidance trajectories explore driving behaviors for a scenario where two pedestrians cross at the same time.}
    \label{fig:3dplan}
\end{figure}

A common approach in the presence of static obstacles is to find \h{a high-level path} using a global planner. \h{This path is} passed to a local motion planner which improves the quality of the plan locally. The majority of existing works \h{following this hierarchy} do not consider dynamic obstacles in their global planner, they delegate dynamic collision avoidance to the local planner directly. Although this is computationally faster, it fails to address that when obstacles move, the set of driving behaviors becomes richer, since the planner needs to decide \textit{when} and \emph{how} to pass the obstacles. Fig.~\ref{fig:3dplan} illustrates these driving behaviors by looking at the planning problem in a \mbox{3-D} state space consisting of \mbox{2-D} position and bounded time. Each driving behavior travels through a distinct section of the state space and can therefore be identified by analyzing their topology in the collision-free space.%

In this work we explicitly consider the possible driving behaviors in a dynamic environment by planning topologically distinct guidance trajectories in the state space. We select the most suitable guidance trajectory based on a high-level cost. This trajectory is passed to a Model Predictive Contouring Controller (MPCC)~\cite{brito_model_2019} that locally optimizes the motion of the robot while following the selected guidance trajectory, resulting in a fast and safe motion plan.
 
\paragraph*{Related Work}
Local optimization can be leveraged to plan locally optimal trajectories. Model Predictive Control (MPC) is often used to optimize planning performance (e.g., speed and comfort) while satisfying constraints (e.g., collision avoidance, vehicle model, actuator limits) offering a flexible and safe framework that can include road following~\cite{schwarting_safe_2018} and dynamic collision avoidance in the deterministic~\cite{brito_model_2019} and uncertain~\cite{zhu_chance-constrained_2019}, \cite{wang_fast_2020}, \cite{de_groot_scenario-based_2021} case. \h{A limitation} of MPC motion planners is that their solution is only guaranteed to be locally optimal, which can lead to unsafe or unexpected driving behavior when the local optimum corresponds to unsuitable (e.g., aggressive or slow) driving behavior.

Global planners such as Rapidly expanding Random Trees (RRT)~\cite{lavalle_rapidly-exploring_1998}, RRT*~\cite{karaman_sampling-based_2011} and Probabilistic RoadMaps (PRM)~\cite{kavraki_probabilistic_1996} in principle resolve this issue, but are typically not fast enough to consider dynamic obstacles, especially when the robot's dynamic constraints are considered.

Topology-based planning methods identify driving behaviors through the environment by comparing trajectory topologies. When two trajectories can be smoothly transformed into each other without colliding with an obstacle, they are \emph{homotopy} equivalent. In~\cite{bes_path_2012}, methods using this measure are divided in three groups. The first group plans in a given homotopy class, for example, road rules are used in~\cite{bender_combinatorial_2015} to motivate a desired homotopy class in highway driving. 

The second group leverages structure in the environment to enumerate possible homotopy classes after which trajectories in a subset of the classes are computed. An example of this approach is~\cite{park_homotopy-based_2015} where a 2-D workspace is decomposed with a trapezoidal decomposition. Similarly, in~\cite{yi_model_2019}, \cite{altche_partitioning_2017} homotopy classes are derived from the road structure. The works~\cite{bhattacharya_persistent_2015}, \cite{bhattacharya_search-based_2010} compute a homology\footnote{Homology differs slightly from homotopy. See~\cite{bhattacharya_persistent_2015} for definitions.} 
invariant that for each obstacle identifies the rotations of a path around it. Graphs extended with this invariant can be used to plan a path in each homology class.

The third group evaluates the homotopy of a trajectory after it is found. In~\cite{rosmann_integrated_2017}, \cite{rosmann_online_2021}, the homology invariant from~\cite{bhattacharya_persistent_2015} is applied to extend a PRM graph around static obstacles and the resulting high-level trajectories are optimized by a local planner. 
An alternative to homotopy, Universal Visibility Deformation (UVD) (based on VD~\cite{jaillet_path_2008}), introduced in~\cite{zhou_robust_2020} efficiently compares the topology of two trajectories. A UVD aware visibility-PRM (see~\cite{simeon_visibility-based_2000}) is presented to plan mulitple distinct trajectories in real-time for drone flight. The same method is leveraged in~\cite{penicka_learning_2022} and~\cite{penicka_minimum-time_2022} to achieve state-of-the-art results for drone flight in static environments.

Previous works~\cite{bender_combinatorial_2015, yi_model_2019, altche_partitioning_2017} in the second group plan among dynamic obstacles in the state space, all of which assume and leverage road structure. 
Works~\cite{zhou_robust_2020, penicka_learning_2022,penicka_minimum-time_2022} in the third group plan drone flight in unstructured \mbox{3-D} environments, but do not consider dynamic obstacles. 

\paragraph*{Contribution}
In this work, we present a method in the third group, based on~\cite{zhou_robust_2020} to plan trajectories through dynamic environments without relying on road structure. The contributions of this work are as follows: 
\begin{enumerate}
    \item A planner that considers multiple topologically distinct high-level trajectories in dynamic environments, without assuming a structured environment. The most suitable trajectory is selected as initialization (guidance) for a local planner.
    \item An algorithm to identify and propagate topology information of trajectories to successive iterations, making the planner behavior stable and consistent.
\end{enumerate}
We validate the proposed planner on a mobile robot both in simulation and in real-world experiments. Our results indicate that the addition of the high-level planner results in faster trajectories, where this improvement increases \h{as the driving scenarios become more crowded}. In addition, we observe less collisions in crowded scenarios. Our planner is implemented in ROS/C++ and will be released open-source.

\section{Problem Formulation}
We model the robot motion by the deterministic discrete-time nonlinear dynamics
\begin{equation}
    \b{x}_{k + 1} = f(\b{x}_k, \b{u}_k),
\end{equation}
where $\b{x}_k\h{\in\mathbb{R}^{n_x}}$ and $\b{u}_k\h{\in\mathbb{R}^{n_u}}$ are the state and input at discrete time instance $k$, \h{$n_x$ and $n_u$ are the state and input dimensions respectively} and the state contains the 2-D position of the robot $\b{p}_k = (x_k, y_k)\in\mathbb{R}^2\h{ \subseteq{\mathbb{R}^{n_x}}}$. 
Obstacles move in the same space as the robot. The position of obstacle $j$ at time $k=0$ is denoted $\b{o}^j_0\in\mathbb{R}^2$ and we assume that for each obstacle predictions of its motion over the next $N$ time steps (i.e., $\b{o}^j_1, \hdots, \b{o}^j_{N}$) are available to the robot at each time instance. We model the obstacles and robot area with a single disc each, with radius $r_{\textrm{obs}}$ and $r$, respectively. We further assume that the robot is given a high-level reference path to follow, e.g., a straight line to the goal. The robot tracks the reference path through the workspace (it can deviate from it) without colliding with the obstacles. 

\section{Global Guidance}\label{sec:method}
The proposed planner consists of two components, the first computes a global guidance trajectory to guide the second component, a local planner, to the global optimum in the dynamic environment. We design the guidance trajectory search to be light-weight and fast, giving an estimate in the vicinity of the global optimum. The local planner is initialized from the guidance trajectory and leverages more accurate robot models and constraints to locally obtain a safe high quality trajectory.



\newcommand\mw{0.21}
\newcommand\ms{35}
\begin{figure*}
    \centering
    \begin{subfigure}[t]{\mw\textwidth}
         \centering
         \includegraphics[width=\textwidth]{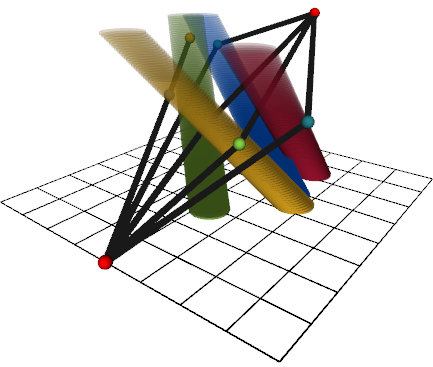}
            \caption{\h{Visibility-PRM graph, with spheres denoting guards (orange), start and goal (red) and connector nodes (colored).}}
         \label{fig:graph}
     \end{subfigure}
     \hspace{\ms pt}
    \begin{subfigure}[t]{\mw\textwidth}
    \centering
         \includegraphics[width=\textwidth]{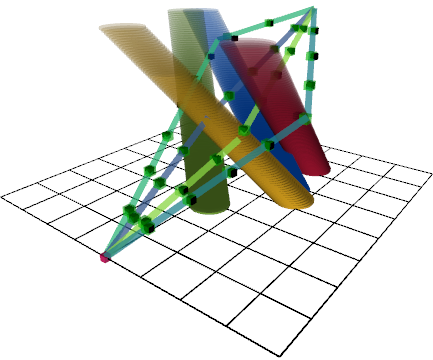}
            \caption{\h{Geometric paths (colored lines) and cubic spline points (cubes) before (green) and after optimization (black).}}
         \label{fig:splines}
     \end{subfigure}
     \hspace{\ms pt}
     \begin{subfigure}[t]{\mw\textwidth}
         \centering
         \includegraphics[width=\textwidth]{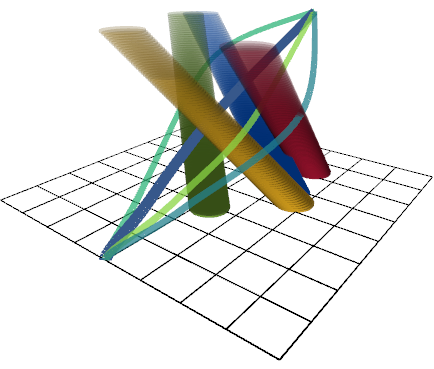}
            \caption{Guidance trajectories with the picked trajectory \h{(dark blue)} thickened.}
         \label{fig:trajectories}
     \end{subfigure}
    \caption{\h{Our high-level guidance method, viewed in the state space $\mathcal{X}$. (a) A graph is constructed using visibility-PRM. (b) Geometric paths are found through graph-search and are sampled and \h{smoothed}.} (c) Cubic splines are fitted through the resulting points. Based on a high-level cost, one guidance trajectory is selected, to be tracked by a local planner.}
    \label{fig:guidance}
\end{figure*}

\subsection{Dynamic Trajectory Planning}
We explicitly consider multiple local optimal driving behaviors in the presence of dynamic obstacles by planning high-level collision-free paths in the state space.
The state space, composed of the workspace and time, is described by $\mathcal{X} = \mathbb{R}^2 \times [0, T]$, with $[0, T]$ a continuous domain. A trajectory is a continuous path through the state space, $\b{\tau} : [0, 1]\rightarrow \mathcal{X}$. The area of the workspace occupied by the union of obstacles at time t is denoted by $\mathcal{O}_t\subset\mathbb{R}^2$ and the obstacle set in the state space is thus $\mathcal{O} := \bigcup_{\forall t\in [0,T]} (\mathcal{O}_t,t) \subset \h{\mathcal{X}}$. Obstacles puncture holes in the state space,  
which make the collision-free space nonconvex and results in the existence of multiple locally optimal trajectories.




The topology of trajectories can be analyzed to identify a single trajectory for each local optimum. To distinguish trajectory topologies, we require a comparison function, 
\begin{equation}
    \mathcal{H}(\b{\tau}_i, \b{\tau}_j, \h{\mathcal{O}})\begin{cases}
        1 & \b{\tau}_i, \b{\tau}_j \textrm{ topologically equivalent},\\
        0 & \textrm{otherwise}.
    \end{cases}
\end{equation}
We then seek to find the set of topologically distinct trajectories $\mathcal{T}^* = \{\b{\tau}_0, \b{\tau}_1, \hdots, \b{\tau}_{N_\tau}\}$ in the workspace, where 
\begin{equation}
    \mathcal{H}(\b{\tau}_i, \b{\tau}_j, \h{\mathcal{O}}) = 0, \ \forall \b{\tau}_i, \b{\tau}_j \in \mathcal{T}^*, \ i \neq j.
\end{equation}
In this work, we adopt the topology measure UVD~\cite{zhou_robust_2020}. 
Two trajectories are in the same UVD class (topology equivalent) if points along the trajectories can be connected, without intersecting with obstacles. 

\begin{definition}{\normalfont{\cite{zhou_robust_2020}}}
    Two trajectories $\b{\tau}_1(s), \b{\tau}_2(s)$ parameterized by $s \in [0, 1]$ and satisfying $\b{\tau}_1(0) = \b{\tau}_2(0)$, $\b{\tau}_1(1) = \b{\tau}_2(1)$, belong to the same uniform visibility deformation class, if for all $s$, line $\overline{\b{\tau}_1(s)\b{\tau}_2(s)}$ is collision-free.
\end{definition}%
\h{In practice, we check collisions for $s$ at discrete intervals along the trajectories.}

\subsection{Visibility-PRM} 
We build on Visibility-PRM~\cite{zhou_robust_2020} to compute a sparse representation of the paths from start to goal, with distinct topologies.
We introduce important adaptations to repurpose the approach in the state space domain, that includes time, to ensure that the algorithm gives consistent outputs over successive time steps. The modified Visibility-PRM algorithm is given in Algorithm~\ref{alg:prm} and is detailed below. 

PRM initializes its graph with start ($\b{x}_0$) and goal ($\b{x}_N$) nodes. States for new nodes are drawn at random from a \textit{feasible} state distribution $\b{x}^{(i)} \sim \mathbb{P}_{\textrm{PRM}}$ (\textbf{NewSample} line \h{$8$}) and if possible, the new node is connected to the graph.

Visibility-PRM distinguishes between \textit{Guard} and \textit{Connector} nodes, as follows, to ensure that the graph is sparse. The start and goal nodes are initialized as guards (lines $1$,~$2$). For each new sample, we find the guards that it can be connected to without colliding (\textbf{VisibleGuards} line \h{$9$}). If it connects to $0$ guards, it is added as \emph{guard} (\textbf{AddGuard} line \h{$11$}). When a sample connects to exactly $2$ guards, it becomes a \emph{connector} (\textbf{InitializeConnector} line~\h{$17$}). For connectors we verify \h{that} the connection is dynamically feasible (\textbf{ConnectionInvalid} lines \h{$13$, $15$}). \h{We then construct a piecewise linear path (\textbf{Path}) between a connector and the guards.} Lines \h{$22$-$29$} replace an existing connector when the newly sampled connector is in the same UVD class and the new path is shorter. New connectors with a distinct UVD class are added by \textbf{AddConnector} in line \h{$32$}.

A depth-first graph search augmented with a visited node list, similar to~\cite{rosmann_integrated_2017}, computes paths on the graph from start to goal, giving the UVD distinct trajectories $\mathcal{T}^*$. We refer to these trajectories as the \emph{geometric trajectories}. An example of the result is visualized in Fig.~\ref{fig:graph}.

{
    \begin{algorithm}[t]
    \caption{Proposed Visibility-PRM}
    \label{alg:prm}
    \KwIn{$\mathcal{O}$, $\b{x}_0$, $\b{x}_N$, Previous nodes $\mathcal{G}^-$}

    \textbf{AddGuard}($\b{x}_0$, $\mathcal{G}$)\\
    \textbf{AddGuard}($\b{x}_N$, $\mathcal{G}$)

    \While{\normalfont{Below sample and time limit}}{
        \h{\normalfont{reintroduce} $\leftarrow$ Not all samples $\mathcal{G}^-$ were reintroduced} \\ 
        \h{\If{\normalfont{reintroduce}}{
            $\b{x}$, $\mathcal{N}^-$ $\leftarrow$ \textbf{ReintroduceSample($\mathcal{G}^-$)}\\
        }\Else{
            $\b{x} \leftarrow$ \textbf{NewSample($\b{x}_0$, $\b{x}_N$)} \\
        }}

        $\mathcal{L} \leftarrow$ \textbf{VisibleGuards}(\h{$\b{x}$, $\mathcal{G}$, $\mathcal{O}$})\\
        \If{$|\mathcal{L}| = 0$ (No guards visible)}{
            \textbf{AddGuard}($\h{\b{x}}$, $\mathcal{G}$)
        }\ElseIf{$|\mathcal{L}| = 2$ (Exactly 2 guards visible)}{

            \If{\normalfont{\textbf{ConnectionInvalid}(\h{$\mathcal{L}_0$, $\b{x}$})}}{
                \textbf{Continue}
            }
            \If{\normalfont{\textbf{ConnectionInvalid}{(\h{$\b{x}$, $\mathcal{L}_1$})}}}{
                \textbf{Continue}
            }

            $\mathcal{N} \leftarrow$ \textbf{InitializeConnector}(\h{$\b{x}$})\\

            \If{\h{\normalfont{reintroduce}}}{
                \h{Transfer segment ID of $\mathcal{N}^-$ to $\mathcal{N}$}
            }

            $\h{\b{\tau}} \leftarrow$ \textbf{Path}(\h{$\mathcal{L}_0, \b{x}, \mathcal{L}_1$})\\

            distinct $\leftarrow$ True\\
            \For{\normalfont{Shared neighbour} \h{$\b{x}^j, \mathcal{N}^j$} \normalfont{of} \h{$\b{x}, \mathcal{N}$} \normalfont{in} $\mathcal{L}$}{
                $\h{\b{\tau}}_j \leftarrow$ \textbf{Path}(\h{$\mathcal{L}_0, \b{x}^j, \mathcal{L}_1$})\\
                \If{$\mathcal{H}(\h{\b{\tau}}, \h{\b{\tau}}_j, \h{\mathcal{O}}) = 1$}{
                    distinct $\leftarrow$ False\\

                    \If{\normalfont{\textbf{Length}}($\h{\b{\tau}}$) $<$ \normalfont{\textbf{Length}}($\h{\b{\tau}}_j$)}{
                        \h{Transfer the segment ID of $\mathcal{N}^j$ to $\mathcal{N}$} \\
                        \h{In $\mathcal{G}$, replace connector $\mathcal{N}^j$ with $\mathcal{N}$}
                    }
                    \textbf{Break}\\
                }
            }

            \If{\normalfont{distinct}}{
                \If{\h{\normalfont{reintroduce} is False}}{
                    \h{Initialize segment ID of $\mathcal{N}$ with unused ID}\\
                }
                \textbf{AddConnector}(\h{$\mathcal{N}$, $\mathcal{G}$})
            }
        }
        
    }
    \KwOut{$\mathcal{G}$}
\end{algorithm}

\subsection{Propagating Guidance Trajectories}
Because the search for guidance trajectories is repeated at each time step, consistency between successive iterations must be guaranteed to prevent the robot from switching between different trajectories. We address this by marking nodes in the graph with a topology identifier, then reintroducing the nodes in the next iteration with their identifiers. This allows us to reidentify and favour the selected guidance trajectory from the previous iteration.


Each segment consisting of a connector and two guards is in a distinct UVD class by construction. We assign to each connector $i$ a \emph{segment ID}, $\alpha_i \in \mathbb{Z}^+$, that uniquely identifies this segment (line \h{$31$}). When a connector is replaced, the ID is transferred (line \h{$27$}). This is illustrated in Fig.~\ref{fig:prm-association}.


\begin{figure}[t]
    \centering
    \begin{subfigure}[t]{0.15\textwidth}
         \centering
         \includegraphics[width=\textwidth]{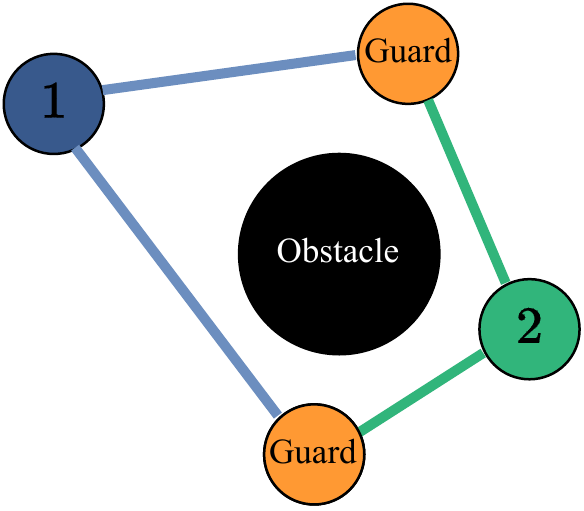}
            \caption{Segments with new topology}
         \label{fig:prm_new}
     \end{subfigure}
    \hspace{10pt}
    \begin{subfigure}[t]{0.15\textwidth}
    \centering
         \includegraphics[width=\textwidth]{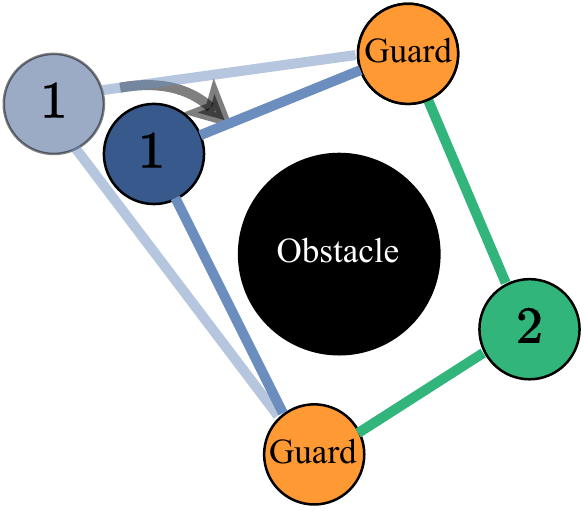}
            \caption{Segments with existing topology}
         \label{fig:prm_replace}
     \end{subfigure}
    \caption{Each segment is associated with a topological ID. (a) Two segments with new distinct topologies are given unique IDs. (b) A new connector creates a shorter segment within an existing topology, the ID is transferred.}
    \label{fig:prm-association}
\end{figure}

Each geometric trajectory is composed of one or more segments. We associate each geometric trajectory $i$ with a \emph{trajectory ID}, $\beta_i \in \mathbb{Z}^+$ and maintain a mapping between each trajectory and its segments, making it possible to reidentify each trajectory in successive iterations. 
For example if trajectory $\b{\tau}_1$ with ID $\beta_i = 1$ contains segments $\alpha_1 = 1, \alpha_2=4$, we first save the mapping $1 \rightarrow \{1, 4\}$. In the next iteration, a trajectory consisting of segments ${1, 4}$ is reassigned the ID $1$. 

In line \h{$6$} of Algorithm~\ref{alg:prm}, \h{\textbf{ReintroduceSample}} reintroduces all nodes of the previous graph ($\mathcal{G}^-$) before sampling new states. Because a sampling time passes between these iterations, the time coordinate of each node $(x, y, t)$ in the previous graph needs to be updated to $(x, y, t-h)$, where $h$ is the sampling time. 
When the time coordinate becomes zero, we resample a new node halfway along the trajectory. 

Since the connectors posses\h{s} a segment ID, the segment associations are carried over from the previous iteration (line \h{$19$}). After new paths are found, we reidentify previous trajectories from their segments. These steps result in consistent identification of UVD distinct trajectories and improve \h{the overall planner performance}, since trajectories and their topology are propagated to successive iterations.

\subsection{Spline Optimization}
The geometric trajectories do not satisfy the kinematic constraints of the robot (they are discontinuous) and therefore cannot be followed by a low-level controller. We smoothen these trajectories in two steps as visualized in Figs.~\ref{fig:splines} and~\ref{fig:trajectories}. The first step samples the trajectories and optimizes the resulting points. The second step fits a smooth curve through the optimized points. 

\paragraph*{\h{Step 1}} We first convert the state space paths to time parameterized trajectories, $\b{\tau}_{geometric} : [0, T] \rightarrow \mathbb{R}^2$. Each trajectory is sampled with regular intervals $\Delta t$ to obtain a set of control points $\b{Q} = [\b{Q}_0, \hdots, \b{Q}_N]$, with $\b{Q}_i \in \mathbb{R}^2$.

We optimize these control points to improve smoothness both in position and velocity. 
The optimization problem is designed to be quadratic and unconstrained to minimize computation times. We define the cost as
\begin{equation}
    J_{\textrm{points}} = J_{\textrm{geo}} + J_{\textrm{smooth}} + J_{\textrm{obst}} + J_{\textrm{vel}},\label{eq:spline_cost}
\end{equation}
where the geometric cost~\cite{zhou_robust_2020} penalizes distance to the \h{original} control points $\bar{\b{Q}}$ on the geometric trajectory,
\begin{equation}
    J_{\textrm{geo}} = \sum_i^N||\b{Q}_i - \bar{\b{Q}}_i||^2.\label{eq:geometric_cost}
\end{equation}
The smoothness cost~\cite{zhou_robust_2020} 
\begin{equation}
    J_{\textrm{smooth}} = \sum_{i=1}^{N-1} ||\b{Q}_{i-1} - 2\b{Q}_i + \b{Q}_{i+1}||^2,
\end{equation}
functions as an elastic band that smoothens the trajectory. 

The obstacle cost is designed to penalize the linear distance $d_{i}^j$ from control point $i$ to the obstacle $j$ by an exponential penalty. To keep the cost quadratic however, we use the second order Taylor expansion. Let $\b{A}_{i, j} = \bar{\b{Q}}_i - \b{o}_k^j$, where time index $k$ matches the time associated with $\b{Q}_i$. \h{Then the cost is given by $J_{\textrm{obst}} = \sum_{i} \sum_{j} \b{Q}_i^T \b{H}_{i, j} \b{Q}_i + \b{f}_{i, j}^T \b{Q}_i$,}
with
\begin{align*}
    \b{H}_{i, j} = \frac{\b{A}_{i, j}\b{A}_{i, j}^T}{2}, \ \b{f}_{i, j} = \h{-(1 + r + \h{r_{\textrm{obs}}})\b{A}_{i, j}} - \h{2}\b{H}_{i, j}\b{o}_k^j.
\end{align*}
The velocity cost penalizes an offset with respect to a tracking velocity $v_{\textrm{ref}}$. We compute this cost by constructing a trajectory that satisfies the velocity tracking cost everywhere, based on the geometric trajectory. We then penalize the distance to that trajectory. The velocity control points of the geometric path can be computed using
\begin{equation}
\b{V}_i = \frac{\b{Q}_{i + 1} - \b{Q}_i}{\Delta t},\ i = 1, \hdots, N-1.
\end{equation}
Then the associated velocity optimized path is given by
\begin{equation}
\b{Q^v_{i+1}} = \b{Q}^v_i + v_{\textrm{ref}} \frac{\b{V}_i}{||\b{V}_i||} \Delta t,
\end{equation}
with $\b{Q}^v_{0}$ equal to the current velocity of the robot. The velocity cost matches \eqref{eq:geometric_cost} \h{with} $\bar{\b{Q}}$ replaced by $\b{Q}^v$.

\paragraph*{\h{Step 2}}Since the optimization problem is quadratic and unconstrained, we obtain the solution in closed-form. We fit cubic splines separately through the optimized $x$ and $y$ position of the control points to obtain a continuous trajectory, consisting of segments $\b{\tau}^i(t) = \begin{bmatrix}\tau^i_x & \tau^i_y\end{bmatrix}^T$, with $\tau^i_x$ (and $\tau^i_y$) given by
\newcommand{\dotr}{\mbox{$\boldsymbol{\cdot}$}}
\begin{align*}
\tau^i_x(t) = a^i_xt^3 + b^i_xt^2 + c^i_xt + d_x,
\end{align*}
which is twice continuously differentiable and passes through the control points. We impose a boundary condition on the initial velocity of the trajectories to ensure that it respects the robot's current velocity. For more details on fitting the cubic splines, we refer to~\cite{kluge_cubic_2021}. The cubic splines together form a smooth trajectory $\b{\tau} : [0, T] \rightarrow \mathbb{R}^2$ from the current robot position to the goal (see Fig.~\ref{fig:trajectories}). \h{Guidance trajectory $\b{\tau}$ is not guaranteed to be collision-free, but is smooth enough to be used by a local planner that enforces collision avoidance.}

\subsection{Spline Selection} 
From the candidate splines generated in each control iteration, we need to select the guidance trajectory that best represents the performance indicators for the robot's motion. 
We consider the following criteria:
\begin{enumerate}
    \item Minimize path length - \h{preferring} short paths.
    \item Minimize difference to \h{preferred} velocity - penalizing too fast or slow driving.
    \item Minimize acceleration - \h{preferring} smooth trajectories.
    \item Consistency - ensuring that when another trajectory improves over the selected trajectory of the previous control iteration, it should significantly outperform it.
\end{enumerate}
We compute the costs for each trajectory by taking samples of positions $\b{p}_i$, velocities $\b{v}_i$ and accelerations $\b{a}_i$ at constant time intervals along the trajectory, resulting in the objective
\begin{equation}
    \begin{aligned}
    J_{\textrm{select}} = &\sum_{i\in\mathcal{I}} w_L||\b{p}_i - \b{p}_{i-1}|| + w_V|| \ ||\b{v}_i|| - \bar{v}||\\
    & + w_a\alpha^{i}||\b{a}_i|| + w_cC,
    \end{aligned}
\end{equation}
where $\mathcal{I}$ is the number of samples, $w$ denotes weights, $\bar{v}$ is the reference velocity, $\alpha \approx 1$ to discount accelerations later in the trajectory and $C$ denotes a constant penalty if this trajectory was \h{not} selected in \h{the} previous iteration. We select the \h{lowest cost trajectory} (thickened in Fig.~\ref{fig:trajectories}).

\section{Local Planning}
The local planner needs to plan a kinematically feasible motion that is collision-free, since this is not guaranteed by the guidance trajectory. 
We introduce the guidance trajectory to the local planner in two ways. First, the solver is initialized with the guidance trajectory. This in itself may not be sufficient to converge to the desired optimum. We therefore also follow the guidance trajectory by using an MPCC~\cite{brito_model_2019} as local planner. This planner is designed to follow the path traced by the guidance trajectory while tracking \h{its} velocity. 

The objective of the MPCC is given by
\begin{equation}
    J_{\textrm{local}} = \sum_{k = 0}^{N_{\textrm{MPCC}}} J_{c, k} + J_{l, k} + J_{v, k} + J_{a, k} + J_{\omega, k},
\end{equation}
where $J_{c, k}$ and $J_{l, k}$ denote the lag and contouring costs as in~\cite{brito_model_2019}, the velocity reference tracking is enforced via $J_{v, k} = ||v_k - \bar{v}_k||$, where $\bar{v}_k$ is the velocity on the guidance trajectory at time step $k$, and the costs $J_{a, k}$ and $J_{\omega, k}$ penalize actuation.

To avoid obstacles one would typically use nonconvex constraints $||\b{p}_k - \b{o}_k|| \geq r + r_{\textrm{obs}}$~\cite{brito_model_2019}. 
However, in practice these may result in switching between local optima, without giving a consistent solution. We employ the linearized version of these constraints, the linear constraint orthogonal to the vector between the robot and obstacle, $\b{A}^T\b{p}_k \leq b$, where
\begin{equation}
\b{A} = \frac{\b{o}_k - \b{p}_k}{||\b{o}_k - \b{p}_k||},\quad b = \b{A}^T(\b{o}_k - A(r + r_{\textrm{obs}})).
\end{equation}
These constraints result in consistent and smooth motion\footnote{In principle we could run a local planner for each guidance trajectory in parallel and choose the best one. This is left for future work.}.


\begin{table}[b]
    \centering
    \caption{Experimental settings. Weights are denoted ``(w)''.}
    \begin{tabular}{|c|c|c|c|c|c|c|}
    \hline \textbf{Overall} & $N_{\textrm{PRM}}$ & $N_{\textrm{MPCC}}$ & $h$ & $\bar{v}$ &  &\\\hline
    & \h{$120$} & $40$ & $0.05$ s & $2$ m/s & &\\\hline
    \textbf{Points (w)} & $J_{\textrm{geo}}$ & $J_{\textrm{smooth}}$ & $J_{\textrm{obst}}$ & $J_{\textrm{vel}}$ & & \\\hline
    & \h{$25$} & $10$ & $0.5$ & \h{$0.01$} & & \\\hline
    \textbf{Splines} & Points & $w_L$ & $w_V$ & $w_a$ & $w_c$ & $\alpha$ \\\hline
    & \h{$20$} & $1$ & $100$ & $100$ & $25$ & $0.95$ \\\hline
    \textbf{MPCC (w)} & $J^{\textrm{MPCC}}_{c, k}$& $J^{\textrm{Guidance}}_{c, k}$ & $J_{l, k}$ & $J_{v, k}$ & $J_{a, k}$ & $J_{\omega, k}$ \\\hline
    & $0.01$ & $0.5$ & $0.01$ & $0.3$ & $0.05$ & $0.05$ \\\hline
    \end{tabular}
    \label{tab:settings}
\end{table}
\section{results}\label{sec:results}
We validate our approach in simulation and real-world experiments, comparing in both cases against a local planner without guidance. Our planner is implemented in ROS/C++ and will be released open-source. A video of the simulations and experiments is available in~\cite{video_de_groot_globally_2023}.

\subsection{Notes on Implementation}
Experimental settings are given in Table~\ref{tab:settings}. The high-level planner is fast enough to plan over a longer horizon than the local planner while remaining real-time (i.e., $N_{\textrm{PRM}} > N_{\textrm{MPCC}}$). In crowded scenarios, this leads to smoother trajectories as the robot can adapt its high-level maneuvre earlier. We select the goal for PRM as the point along the reference path reached when the robot drives at the \h{preferred} velocity. It is projected to the nearest collision-free position if necessary. When the goal cannot be reached, we reduce the horizon. PRM nodes are sampled (i.e., $\mathbb{P}_{\textrm{PRM}}$) in a forward directed arc considering velocity and acceleration limits.

\newcommand\sw{0.19}
\newcommand\ssp{4}
\begin{figure*}
    \centering
    \begin{subfigure}{\sw\textwidth}
        \centering
        \includegraphics[width=\textwidth]{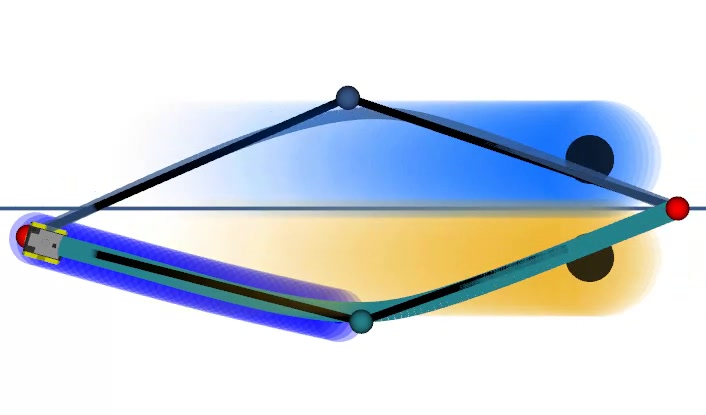}
        \caption{Head-on $2$}
    \end{subfigure}
    \hspace{\ssp pt}
    \begin{subfigure}{\sw\textwidth}
        \centering
        \includegraphics[width=\textwidth]{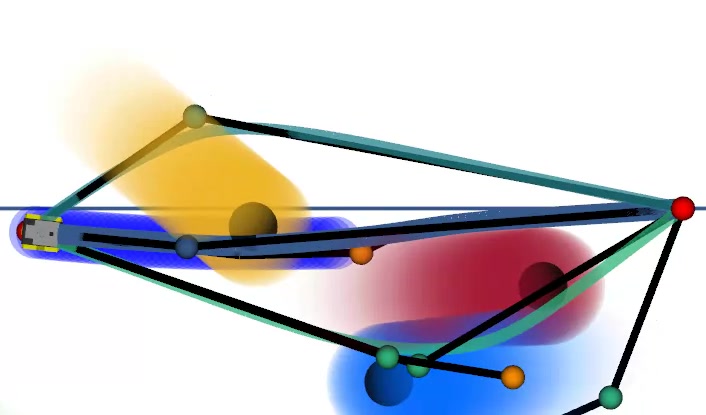}
        \caption{Random $4$}
    \end{subfigure}
    \hspace{\ssp pt}
    \begin{subfigure}{\sw\textwidth}
        \centering
        \includegraphics[width=\textwidth]{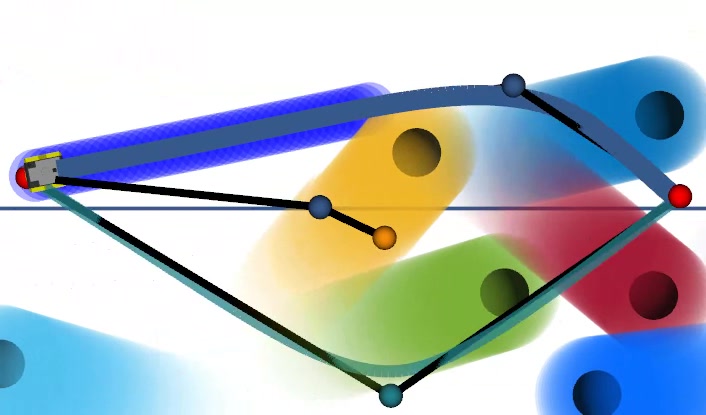}
        \caption{Random $8$}
    \end{subfigure}
    \hspace{\ssp pt}
    \begin{subfigure}{\sw\textwidth}
        \centering
        \includegraphics[width=\textwidth]{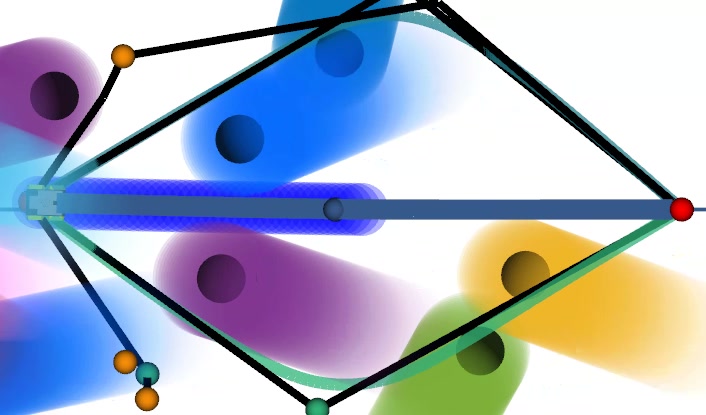} 
        \caption{Random $16$}
    \end{subfigure}
    \caption{Snapshots of the simulations, viewed in the state-space. Pedestrians (black discs) are visualized with their predicted area (colored disc) inflated by the robot area. Visualization of the graph and guidance trajectories are identical to Fig.~\ref{fig:guidance}. The robot's local motion plan is denoted by blue discs, indicating the predicted area occupied by the robot.}
    \label{fig:snapshots}
\end{figure*}
\begin{table*}
    \centering
    \caption{\h{Statistical results for the task duration, number of collisions and computation times when comparing MPCC with and without guidance over \h{$200$} simulations each in $4$ different scenarios. \h{Results are denoted as ``avg (std)'' over experiments except for collisions. We denote with ``High-Level Planning'' the time spent to compute the guidance trajectory.}}}
    \begin{tabular}{|l|l|c|c|c|c|}
        \hline \textbf{Scenario} & \textbf{Method}  &\textbf{Task Duration [s]} & \textbf{Collisions} & \h{\textbf{Computation Time [ms]}} & \h{\textbf{Computation Time High-Level Planning [ms]}}\\\hline
        \multirow{2}{*}{Headon 2} & LMPCC & 13.3 (0.3) & 160 & 17.1 (8.9) & -\\
        &Guidance MPCC & \textbf{12.3} (0.1) & \textbf{0} & \textbf{16.5} (2.9) & 6.5 (2.1) \\\hline
        \multirow{2}{*}{Random 4} & LMPCC & 12.3 (0.3) & \textbf{0} & 17.2 (3.0) & - \\
        &Guidance MPCC & \textbf{12.2} (0.2) & 1 & \textbf{16.2} (2.6) & 6.0 (1.9) \\\hline
        \multirow{2}{*}{Random 8} & LMPCC & 12.7 (0.6) & 3 & \textbf{17.2} (3.1) & -\\
&Guidance MPCC & \textbf{12.4} (0.4) & 3 & 17.6 (3.7) & 6.5 (2.3) \\\hline
\multirow{2}{*}{Random 16} & LMPCC & 13.5 (1.1) & \textbf{16} & \textbf{19.4} (4.3) & -\\
&Guidance MPCC & \textbf{13.2} (1.0) & 17 & 22.1 (6.7) & 8.5 (4.6) \\\hline
    \end{tabular}
    \label{tab:simulation_results}%
\end{table*}

\begin{figure}[t]
    \centering
    \hspace{8pt}
    \begin{subfigure}[t]{0.38\textwidth}
        \centering
        \includegraphics[width=\textwidth,trim={0cm 0.1cm 0cm 0.6cm},clip]{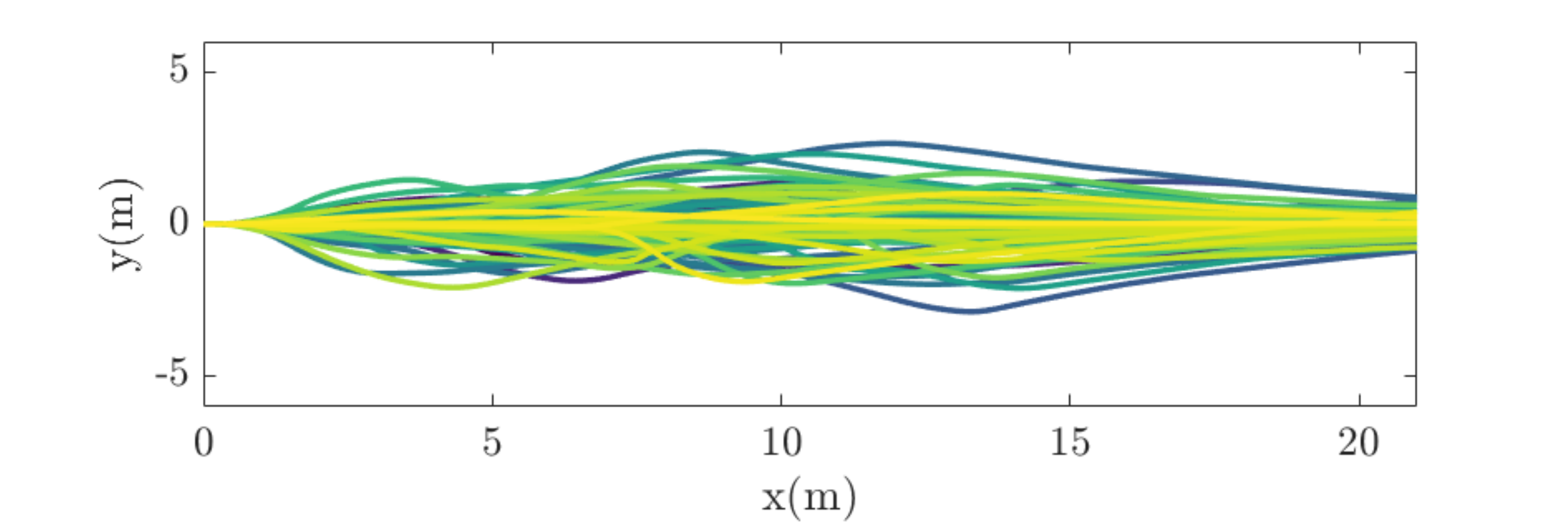}
        \caption{Guidance-MPCC}
        \label{fig:traj_guided}
    \end{subfigure}\newline
    \begin{subfigure}[t]{0.38\textwidth}
        \centering
        \includegraphics[width=\textwidth,trim={0cm 0.1cm 0cm 0.6cm},clip]{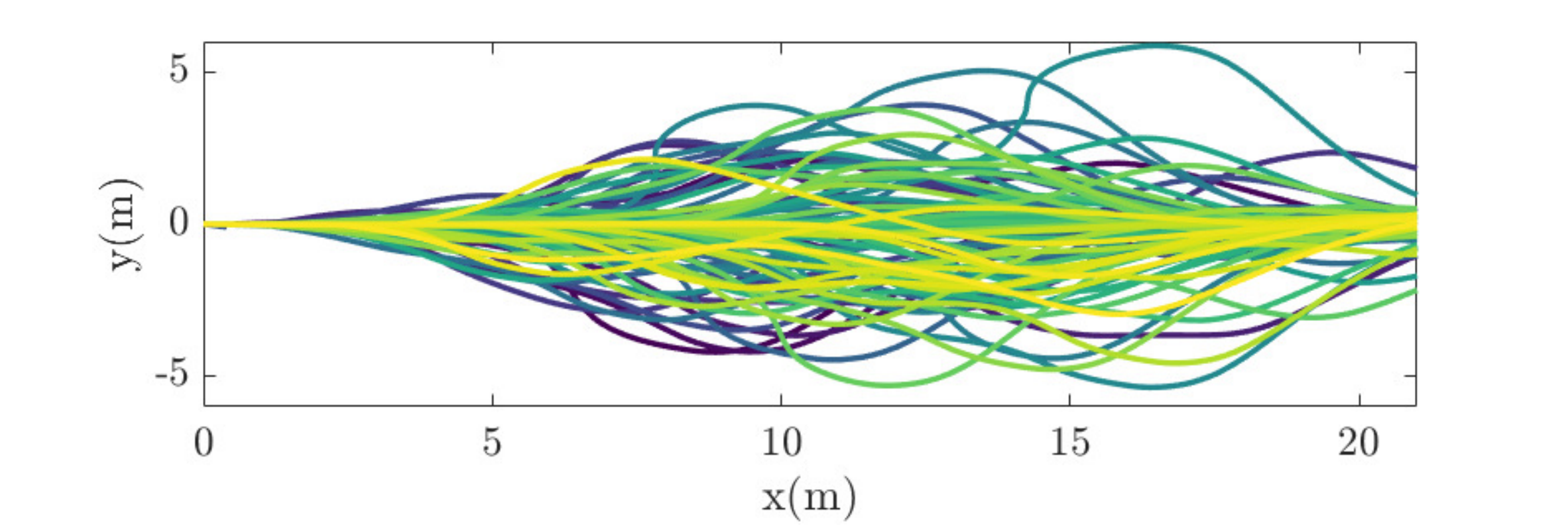}
        \caption{MPCC}
        \label{fig:traj_regular}
     \end{subfigure}
    \caption{A comparison of trajectories with and without guidance in a randomized environment with $8$ pedestrians.}
    \label{fig:trajectories-8}
\end{figure}

\subsection{Simulated Guidance Ablation Study}
Our simulations consider four environments with pedestrians. The first environment (\emph{head-on}) consists of a straight road with two pedestrians moving towards the robot. The other environments contain $4, 8$ and $16$ pedestrians with random start positions and velocities near the reference path. \h{The random scenarios are identical for all planners (i.e., we use the same random seed)}. 
In all simulations, pedestrians move with constant velocity. We reset the simulation when the robot reaches the end of the road in the x-direction and repeat each experiment \h{$200$} times.

Statistical results are given in Table~\ref{tab:simulation_results} and snapshots are shown in Fig.~\ref{fig:snapshots}. The guidance allows the robot to consistently navigate around the pedestrians in the head-on scenario. Without guidance, indecisiveness of the planner leads to infeasibility and collisions in most simulations.
In the randomized environments, the guidance reduces the task duration by \h{$2, 3$ and $5\%$}, with a larger improvement in more crowded environments. Trajectories for the $8$ pedestrian case, visualized in Fig.~\ref{fig:trajectories-8}, show that guidance allows the planner to choose faster driving behaviors. \h{Guidance MPCC collides slightly more often than the baseline.} When obstacles block the path, the high-level planner may \h{not} find a guidance trajectory. \h{We then use} the last computed guidance trajectory\h{, which can lead to collisions. Future work may resolve this problem by considering multiple goals.}

\begin{figure}[t]
    \centering
    \begin{subfigure}[b]{0.15\textwidth}
        \centering
        \includegraphics[clip,width=\textwidth]{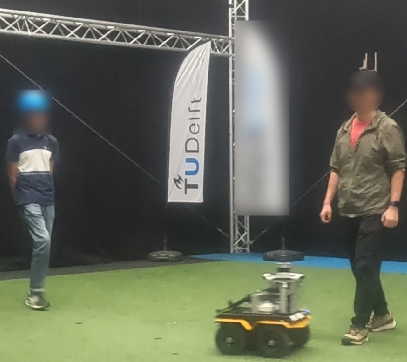} 
        \caption{Setup}
    \end{subfigure}
    \begin{subfigure}[b]{0.15\textwidth}
        \centering
        \includegraphics[trim={2cm 0 1.5cm 0},clip,width=\textwidth,angle=90,origin=c]{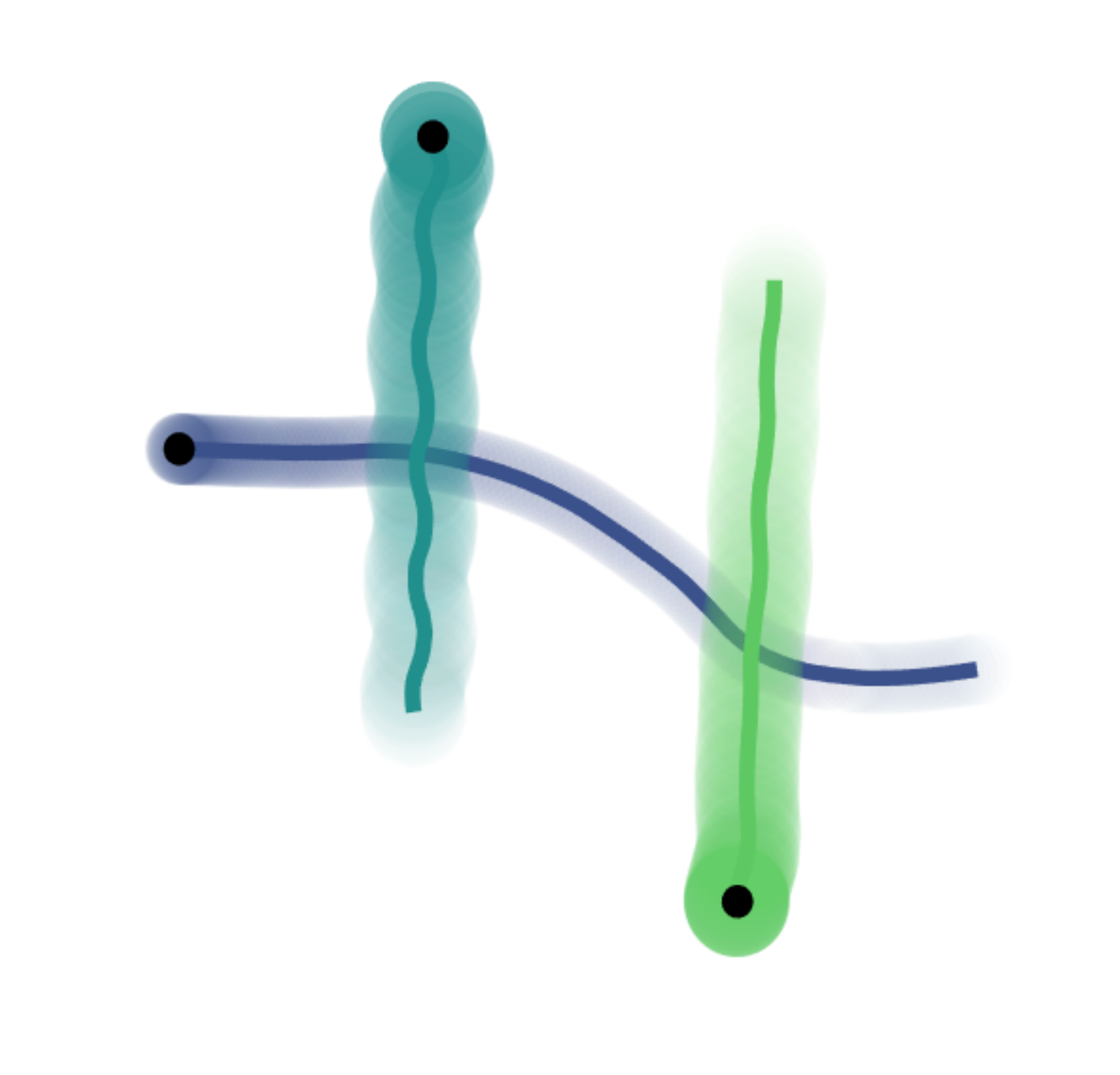} 
        \caption{Crossing Case}
    \end{subfigure}
    \begin{subfigure}[b]{0.15\textwidth}
        \centering
        \includegraphics[trim={2cm 0 1.5cm 0},clip,width=\textwidth,angle=90,origin=c]{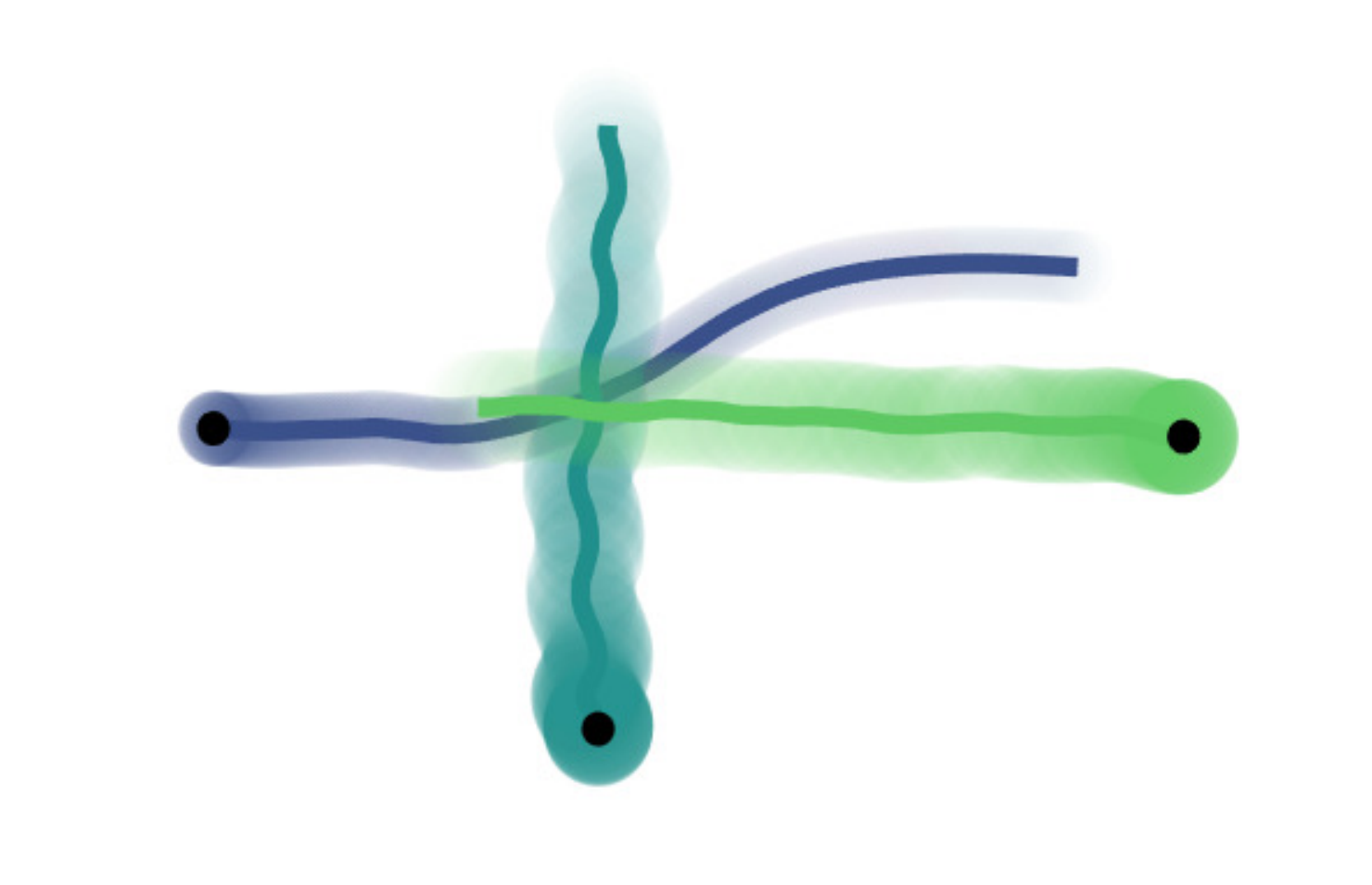} 
        \caption{Mixed Case}
    \end{subfigure}
    \caption{Trajectories recorded in the real-world experiments for the robot (dark blue) and pedestrians (green). Start positions are denoted in black.}
    \label{fig:experiment_trajectories}
\end{figure}

\subsection{Real-World Experiments}
We \h{deploy} the proposed planner experimentally on a mobile robot (Clearpath Jackal) navigating among pedestrians. The robot is equipped with an Intel i5 CPU@2.6GHz. Localization of the robot and pedestrians is obtained from a marker based tracking system and pedestrian predictions assume constant velocity, where the velocity is obtained from Kalman filtered position data. 

Trajectories for $2$ scenarios and the setup are visualized in Fig.~\ref{fig:experiment_trajectories}. In the first scenario, the guidance allows the robot to pass behind the last pedestrian. In the second scenario, the robot moves left to evade both pedestrians efficiently. 

\section{Conclusion}
This work presented a novel planner for autonomous navigation in dynamic environments. The planner finds \h{distinct} high-level trajectories to guide a local optimization-based planner to a global optimal plan. Guidance trajectories are computed and tracked over successive \h{control} iterations. 

We showed that \h{the resulting planner leads to shorter average task duration times than} 
the local planner in isolation, with larger improvement in crowded environments. Real-world experiments further validated the proposed approach.

Further research could explore applications of the guidance trajectory search to predict human motion or to endow the local planner with socially compliant motion.

\bibliographystyle{IEEEtran}
\bibliography{references_zotero}

\end{document}